\documentclass[10pt,conference]{IEEEtran}
\IEEEoverridecommandlockouts
\usepackage{cite}
\usepackage{amsmath,amssymb,amsfonts}
\usepackage{graphicx}
\usepackage{textcomp}
\usepackage{xcolor}
\usepackage[hidelinks]{hyperref}
\usepackage{mdframed}
\usepackage{tabularray}
\usepackage{tikz}
\newcommand*\circled[1]{\tikz[baseline=(char.base)]{
            \node[shape=circle,draw,fill=black,text=white,inner sep=1pt] (char) {#1};}}
\usepackage{multirow}
\def\BibTeX{{\rm B\kern-.05em{\sc i\kern-.025em b}\kern-.08em
    T\kern-.1667em\lower.7ex\hbox{E}\kern-.125emX}}

\usepackage{algorithm,algpseudocode}
\newcommand{\algmargin}{\the\ALG@thistlm}
\makeatother
\algnewcommand{\parState}[1]{\State%
    \parbox[t]{\dimexpr\linewidth-\algmargin}{\strut\hangindent=\algorithmicindent \hangafter=1 #1\strut}}

\usepackage{wrapfig}
\usepackage{enumitem}
    
\begin{document}

 \title{Beyond Keywords: A Context-based Hybrid Approach to Mining Ethical Concern-related App Reviews}

\author{
    \IEEEauthorblockN{Aakash Sorathiya, Gouri Ginde}
    \IEEEauthorblockA{\textit{Department of Electrical and Software Engineering} \\
        \textit{University of Calgary}\\
        Calgary, Canada \\
        \{aakash.sorathiya, gouri.ginde\}@ucalgary.ca}
}
\maketitle
\begin{abstract}
With the increasing proliferation of mobile applications in our everyday experiences, the concerns surrounding ethics have surged significantly. Users generally communicate their feedback, report issues, and suggest new functionalities in application (app) reviews, frequently emphasizing safety, privacy, and accountability concerns. Incorporating these reviews is essential to developing successful products. However, app reviews related to ethical concerns generally use domain-specific language and are expressed using a more varied vocabulary. Thus making automated ethical concern-related app review extraction a challenging and time-consuming effort.

This study proposes a novel Natural Language Processing (NLP) based approach that combines Natural Language Inference (NLI), which provides a deep comprehension of language nuances, and a decoder-only (LLaMA-like) Large Language Model (LLM) to extract ethical concern-related app reviews at scale.  Utilizing 43,647 app reviews from the mental health domain, the proposed methodology 1) Evaluates four NLI models to extract potential privacy reviews and compares the results of domain-specific privacy hypotheses with generic privacy hypotheses; 2) Evaluates four LLMs for classifying app reviews to privacy concerns; and 3) Uses the best NLI and LLM models further to extract new privacy reviews from the dataset. Results show that the DeBERTa-v3-base-mnli-fever-anli NLI model with domain-specific hypotheses yields the best performance, and Llama3.1-8B-Instruct LLM performs best in the classification of app reviews. Then, using NLI+LLM, an additional 1,008 new privacy-related reviews were extracted that were not identified through the keyword-based approach in previous research, thus demonstrating the effectiveness of the proposed approach.

\end{abstract}


\section{Introduction}

Mobile applications are created with specific user goals in focus \cite{ebrahimi2022unsupervised}. A user goal can be defined as any conceptual aim the given system should fulfill \cite{lamsweerde2009requirements}. For instance, Sharing Economy applications (like Uber and Airbnb) aim to enhance social capital and stimulate economic development in resource-limited areas \cite{martin2016sharing}. In contrast, the goal of Health\&Fitness applications is to encourage healthy habits among both children and adults \cite{dennison2013opportunities}. However, due to intense market rivalry, the app development cycle often aims to produce functional applications within brief intervals (such as days or weeks), leading developers to stray from their initial objectives frequently. These divergences frequently bring forth ethical concerns such as declining mental health, bias, privacy violations, and manipulation \cite{zuboff2019age, conger_eating_2021, hill_wrongfully_2020, gillespie_are_2019}. Applications that fail to sufficiently consider their users' ethical concerns are often labeled as untrustworthy or even deserted by their users \cite{haggag2022large}. Thus, for applications to endure the market's scrutiny, developers continuously keep track of user feedback through ratings and reviews found in app marketplaces (like Google Play Store). They typically analyze user feedback to gather insights on bug reports, feature suggestions, connectivity issues, resource consumption challenges (e.g., battery life), and interface problems \cite{ciurumelea2017analyzing, di2017surf, li2018mobile, palomba2018crowdsourcing}.

Numerous studies have investigated user perspectives on ethical concerns within software applications. Research conducted by Besmer et al. \cite{besmer2020investigating} and Nema et al. \cite{nema2022analyzing} underscores users' concerns regarding privacy breaches and data security measures in mobile applications. The emergence of discriminatory algorithms and the potential for bias in software functionalities are also significant areas of concern, as highlighted by the findings of Tushev et al. \cite{tushev2020digital} and Olson et al. \cite{olson2023along}. Furthermore, manipulative design tactics that coerce users or take advantage of psychological weaknesses are increasingly worrisome, as noted by Olson et al. \cite{olson2024best}. However, these investigations largely depend on keyword-based sampling from app reviews, which limits the ethical issues users address to a predetermined set of terms.

To overcome this limitation, Harkous et al. \cite{harkous2022hark} suggest using the NLI method. However, they rely on a set of generic privacy hypotheses (derived from generic privacy concepts) overlooking the fact that users' ethical concerns are domain-dependent \cite{ebrahimi2022unsupervised}. For instance, individuals using ridesharing services (e.g., Uber and Lyft) may raise concerns about the constant tracking of their location, while those utilizing financial platforms (e.g., Robinhood and Coinbase) might express concerns regarding the sharing of their social security or banking details with the application. Additionally, NLI with generic hypotheses identifies a high number of false positives (FP) that require further manual analysis to identify ethical concern-related reviews \cite{harkous2022hark}.

\textbf{To address these challenges, in this paper, we propose a} novel Natural Language Processing (NLP) based hybrid approach that combines Natural Language Inference (NLI) and a decoder-only Large Language Model (LLM) to mine ethical concern-related app reviews at scale. We use NLI with domain-specific hypotheses to determine potential ethical concern-related reviews and further process these reviews using LLMs to extract ethical concern-related app reviews.


\textbf{The main contributions of this study can be summarized as follows.} 
\begin{itemize}[leftmargin=*]
    \item To the best of our knowledge, this is the first hybrid approach that utilizes NLI and LLM along with domain-specific privacy hypotheses to extract ethical concern-related app reviews. NLI+LLM demonstrated better results compared to generic privacy hypotheses utilized by Harkous et al. \cite{harkous2022hark}.
    \item We develop domain-specific hypotheses based on the Mental Health (domain-specific) privacy concepts provided by Iwaya et al. \cite{iwaya2023privacy}.
    \item We demonstrate that our proposed hybrid approach (NLI+LLM) can extract concern-related reviews that do not contain predefined wordings used in the keyword-based method in Ebrahimi et al \cite{ebrahimi2022unsupervised}.
    \item \textbf{We open source our source code and dataset\footnote{\url{https://github.com/AakashSorathiya/CHyMER}}} of 1,008 privacy-related reviews (results from our study) that remained unidentified by the previous Ebrahimi et al's \cite{ebrahimi2022unsupervised} study which used a keyword-based approach.
\end{itemize}

The rest of the paper is organized as follows. To determine the research gaps, Section \ref{rw} discusses related work. Section \ref{motivation} presents the motivation for our research through examples. We define our research questions (RQs) and explain preliminaries in Section \ref{rq} and \ref{prem}, respectively. In Section \ref{data}, we describe the dataset and explain our methodology in Section \ref{std}. Section \ref{str} shows and discusses the results of our investigation. Section \ref{threats} lists various threats to the validity of our investigation and Section \ref{conclude} presents concluding remarks and future directions.

\section{Related work} 
\label{rw}
\textbf{App Reviews:} Numerous scholarly studies have assessed the significance of user feedback within app reviews \cite{ciurumelea2017analyzing, pagano2013user, li2018mobile, khalid2014mobile}. Noteworthy contributions from researchers such as Pagano et al. \cite{pagano2013user} and Khalid et al. \cite{khalid2014mobile} have explored app review classification comprehensively. However, these classifications are rather abstract, encompassing categories such as ``commendation", ``utility", ``issue reporting", and ``feature suggestion" \cite{pagano2013user}, and ``operational failure", ``compatibility" and ``user interface design" \cite{khalid2014mobile}.

Furthermore, an investigation conducted by Lu and Liang \cite{lu2017automatic}, utilizing the categorizations established by the International Organization for Standardization (ISO), delineated six distinct review types based on their thematic focus: Usability, Reliability, Portability, Performance, Feature Request (denoting ``Capabilities that a system/product ought to possess"). Additional research also explores trends and implications within the app review landscape, providing insights into user behavior and app store dynamics \cite{martin2016survey}. App reviews can also pinpoint informative reviews for developers \cite{chen2014ar}, and assist in strategizing release planning based on user sentiment \cite{villarroel2016release}. Detailed sentiment analysis of app reviews equips developers with an understanding of specific feature perceptions, thereby guiding future development decisions \cite{gu2015parts}. Moreover, app reviews can facilitate comprehension of user requirements, highlight desired functionalities \cite{iacob2013retrieving}, and inform processes related to software requirements engineering \cite{carreno2013analysis}.

Sorathiya et al's literature review \cite{sorathiya2024ethical} highlights there is still little research that focuses on the ethical concerns mentioned in app reviews and most of these studies focus on single ethical concerns like privacy \cite{harkous2022hark}, accessibility \cite{alomar2021finding}, and discrimination \cite{tushev2020digital}. Besides these studies, previous work studied multiple ethical concerns mentioned in app reviews \cite{olson2024best} and on Reddit \cite{olson2023along}. While the former proposed an initial taxonomy for ethical concerns and applied machine learning (ML) and deep learning (DL) techniques for its classification; the latter focused exclusively on concerns expressed by marginalized communities.

One major drawback with most of these studies is that they use a keyword-based approach for identifying potential concern-related reviews \cite{sorathiya2024ethical}. Only one study: Harkous et al.\cite{harkous2022hark} leveraged NLI for this task and showed the limitations of using a keyword-based approach with a pre-defined set of keywords. However, their approach is based on a set of hand-crafted generic privacy hypotheses, which once again is a limitation since users' ethical concerns are domain-dependent \cite{ebrahimi2022unsupervised}. Additionally, NLI flags a high number of FP which requires a large amount of manual work to extract relevant reviews \cite{harkous2022hark}. To overcome these limitations, in this paper, we utilize domain-specific privacy hypotheses to create a set of privacy hypotheses for NLI to identify potential ethical concern-related app reviews. \\ \\
\noindent \textbf{Large Language Models (LLMs):}
LLMs are categorized into three groups based on their architecture structure: 1) encoder-only LLMs, (Eg: BERT) 2) encoder-decoder LLMs (Eg: RoBERTA), and 3) decoder-only LLMs (Eg: LLaMA) \cite{hou2023large}. Encoder-only LLMs only use the encoder to encode the sentence and understand the relationships between words. The common training paradigm for these models is to predict the mask words in an input sentence \cite{hou2023large}. Encoder-decoder LLMs adopt both the encoder and decoder module. The encoder module is responsible for encoding the input sentence into a hidden space, and the decoder is used to generate the target output text \cite{hou2023large}. Decoder-only LLMs only adopt the decoder module to generate target output text. The training paradigm for these models is to predict the next word in the sentence \cite{hou2023large}.

Recently, LLMs have been widely utilized, due to their ability to solve various problems in the domain of software engineering (SE), where they are currently employed in a multitude of applications, such as testing, code generation, and code summarization \cite{hou2023large}. Historically, conventional SE tasks associated with the examination of natural language have been predominantly approached through the use of encoder-only LLMs, such as BERT \cite{devlin2018bert} along with its derivatives \cite{lan2019albert}, which also incorporate SE-specific enhancements such as Code-BERT \cite{feng2020codebert} and BERTOverflow \cite{tabassum2020code}. Furthermore, the exploration of encoder-decoder LLMs has been widely explored by models such as T5 \cite{raffel2020exploring} and CodeT5 \cite{wang2021codet5} in SE tasks. Additionally, encoder-only and encoder-decoder models have been widely used for the task of app review classification to ethical concerns \cite{sorathiya2024ethical}.

More recently, commencing in 2023, there has been a notable emergence of decoder-only models, including LLaMA \cite{touvron2023llama} and GPT \cite{achiam2023gpt}, which have gained significant traction within the realm of SE research \cite{hou2023large}. These models have been employed in the SE domain for a variety of tasks, including program repair \cite{cao2023study}, code summarization \cite{zhang2020retrieval}, software testing \cite{wang2021well}, natural language translation to code \cite{zan2022large}, code clone detection \cite{dou2023towards}, and code comprehension \cite{yuan2023evaluating}. Additionally, these models require minimal fine-tuning and can produce syntactically and functionally relevant output \cite{hou2023large}. Despite such encouraging outcomes of decoder-only (LLaMA-like) LLMs for various SE tasks \cite{hou2023large}, to the best of our knowledge, LLaMA-like models have not been leveraged in the context of ethical concern-related review extraction yet.

\section{Motivation} \label{motivation}
Extracting ethical concerns-related reviews through manual inspection is a laborious task since app reviews for any mobile app appear in large numbers. Conversely, recent advances in automated requirements extraction rely solely on keyword matching techniques utilizing ML machine learning (ML) and deep learning (DL) methodologies \cite{olson2024best}. The drawback of keyword matching techniques is, that the set of keywords associated with ethical concerns is curated based on a set of pre-identified generic (context-independent) keywords associated with ethical concerns. Although this technique appears to be more efficient than the manual alternative, there are several limitations to this method: the keyword-matching technique fails to account for the fact that keywords designated for specific ethical concerns may not align with the terminology utilized by users in their reviews \cite{alomar2021finding}. Such discrepancies may arise, for instance, from typographical errors made by users. Additionally, the mere occurrence of certain keywords within a review does not inherently imply that the review addresses any ethical concern. For instance, consider the following review extracted from the dataset compiled by Ebrahimi et al. \cite{ebrahimi2022unsupervised}:

\begin{quote}
    \textit{``...I paid 189\$ for 1 month of couples therapy. she then provided me a link for my husband to join us in the consult private room. the first link did not work at all. the second one she provided took him to a different consultant..."}
\end{quote}

This review contains the term ``private", which was considered in the original compilation of keywords to delineate reviews pertinent to privacy concerns \cite{ebrahimi2022unsupervised}. However, in this context, the term ``private" pertains to the private consultation room and is not associated with privacy-related concerns. Consequently, the identification of reviews concerning privacy issues is significantly dependent on contextual interpretation; thus, merely conducting searches for related keywords within the review text might not be an effective approach.

Harkous et al. \cite{harkous2022hark} employed the NLI task \cite{maccartney2009extended} to mitigate the constraints associated with keyword-based search methodologies. They performed an extensive investigation into the privacy concerns articulated by users in app reviews, leveraging the concepts defined in the established privacy taxonomies \cite{solove2005taxonomy, wang2009privacy}. Utilizing these concepts, they formulated 31 privacy hypotheses, which were subsequently applied to the NLI task, aiming for comprehensive coverage of various dimensions within the privacy domain, irrespective of the linguistic variations present in app reviews. Despite addressing the limitations of keyword-based search, this methodology solely relies on generic privacy concepts rather than domain-specific privacy frameworks. Thus, overlooks the fact that users tend to articulate their ethical concerns using a more varied language, unlike specific terminologies, generally used while mentioning concerns related to technical aspects of the application\cite{mcilroy2016analyzing}. 

For instance, consider the following three reviews selected from the domains of MH, finance, and food delivery applications derived from the dataset \cite{ebrahimi2022unsupervised}. The term \textit{Facebook} signifies a privacy-related concern within the MH domain. Conversely, in the food delivery context, the same term denotes a customer support issue, while in the finance sector, it pertains to a user registration concern.

\begin{mdframed}
\textbf{Mental Health}: ``Won’t even let me sign up after collecting all of my Facebook data, just stole my identity." \\
\textbf{Finance}: ``I got zero response back. I even blasted their Facebook but got nothing." \\
\textbf{Food Delivery}: ``It doesn’t recognize my facebook account so I can’t even register for this."
\end{mdframed}

In addition, there are a variety of app domains, each with a set of particular requirements \cite{dragoni2019unsupervised} that collect different types of data. For example, data in the MH domain involves sensitive personal information such as emotional states, therapy progress, and medical history \cite{balcombe2021digital}, while the finance domain handles financial/investment-related data and transactions \cite{widjaja2024privacy}. Consider the following two reviews selected from the domains of MH and finance applications from the data set \cite{ebrahimi2022unsupervised}. The MH app review expresses concern regarding private medical data being linked with Facebook whereas the investing app review highlights the concern regarding confidential banking details being collected.

\begin{mdframed}
\textbf{Mental Health}: ``You have to have a facebook account that steals all of our information including our medical registries" \\
\textbf{Finance}: ``App asked for my bank login to verify the account. did not offer any other solution. i'm not giving my login info to a third party, so i'll just put my money in webull."
\end{mdframed}

Another limitation of using NLI with generic hypotheses is that it identifies a high number of FP that requires further manual analysis to identify relevant reviews related to ethical concerns \cite{harkous2022hark}.

\textbf{Motivated by these limitations of the existing studies}, \textbf{in this paper, we propose }a novel approach for extracting ethical concern-related app reviews. We first address the limitation of NLI by defining a new set of hypotheses derived from the domain-specific privacy taxonomies, and then to reduce the manual work to identify relevant reviews we leverage LLaMA-like LLMs. We utilized Ebrahim et al.'s dataset \cite{ebrahimi2022unsupervised} for this study and extracted new privacy-related app reviews using NLI+LLM.

\section{Research questions (RQs)} \label{rq}
Our three RQs are as follows:

\textbf{RQ1. To what extent can NLI accurately identify potential ethical concern-related app reviews?}

We aim to investigate whether we can use NLI with domain-specific privacy hypotheses to flag the potential concern-related app reviews. These reviews can contain FP, but leveraging NLI filters the large set of unrelated app reviews. NLI has already been utilized by Harkous et al.\cite{harkous2022hark} for identifying potential reviews but our purpose is to show that domain-specific hypotheses yield better results than generic hypotheses used by \cite{harkous2022hark}.

\textbf{RQ2. To what extent we can leverage LLaMA-like LLMs to classify ethical concern-related app reviews?}

NLI identifies a high number of FP which necessitates further manual analysis to identify relevant ethical concern-related reviews \cite{harkous2022hark}. To reduce this manual effort we aim to investigate the efficiency of leveraging LLaMA-like LLMs for identifying relevant reviews. LLaMA-like LLMs have been employed in the SE domain for a variety of tasks and have shown encouraging results \cite{hou2023large}. 

\textbf{RQ3. How effective is our approach in identifying ethical concern-related reviews as compared to the keyword-based approach?}

After evaluating NLI and LLM individually, we select the best-performing models and compare our hybrid approach with the keyword-based approach. We aim to evaluate NLI+LLM for extracting ethical concern-related reviews that do not contain predefined wordings used in the keyword-based method. We utilize the dataset from the previous study \cite{ebrahimi2022unsupervised} and extract new concern-related reviews that were missed by that study based on the keyword approach.

\section{Preliminaries} \label{prem}

\textbf{Natural Language Inference (NLI):}
NLI pertains to the problem of ascertaining whether a natural language hypothesis can logically be derived from a specified premise \cite{maccartney2009extended}. An NLI model is required to evaluate whether a hypothesis is true (i.e. entailment), false (i.e., contradiction), or undetermined (i.e., neutral) in relation to a given premise. For instance, consider a premise stating, ``…collecting all of my Facebook data, just stole my identity…". A hypothesis asserting, ``too much personal data is collected" would be assigned an \textit{entailment} label. Conversely, a hypothesis claiming ``user likes that data privacy is provided" would be designated a \textit{contradiction} label, and a hypothesis positing ``app has a good interface" would be assigned a \textit{neutral} label.

Moreover, this methodology mitigates the dependency on specific keywords due to the extensive linguistic variability present in the premises associated with the hypotheses. For instance, both of the following reviews receive an \textit{entailment} label for the hypothesis ``The user is not aware of how and why their data is being collected, processed, stored, and shared.":

\begin{itemize}
    \item ``Don t bait people in to take their information and sell it and add them to your mailing list" (P(entailment)=0.76)
    \item ``This app has data trackers don t trust any app with your wellbeing that is sending your behavior data to multiple third parties" (P(entailment)=0.87)
\end{itemize}

\textbf{Note that} no review has any words in common with hypotheses, but both of them discuss the concern related to data collection and sharing. Here, P(entailment) denotes the probability of the \textit{entailment} label and is referred to as \textit{entailment\_score}. We use these scores to filter out the potential reviews based on the defined heuristics. \\ \\
\textbf{Large Language Models (LLMs):} LLMs based on the transformer architecture \cite{vaswani2017attention} have introduced a significant advancement in the field of NLP \cite{devlin2018bert, zhao2023survey}. LLaMA-like LLMs have demonstrated the power and versatility of the transformer architecture when scaling up the number of parameters \cite{kaplan2020scaling}. In particular, they exhibit emergent abilities that arise suddenly at large scales and cannot be extrapolated from smaller models. The mechanisms behind emergence are not fully understood, but hypothesized factors include model capacity, depth, and ability to leverage huge amounts of pre-training data \cite{wei2022emergent}.

Many of those models, after their pre-training phase, are further trained to follow instructions through Reinforcement Learning for Human Feedback (RLHF) \cite{ouyang2022training}, a technique for training models to align with human goals by providing feedback in the form of rewards \cite{christiano2017deep}. This additional fine-tuning makes them a better choice for many NLP tasks because pre-trained models are excellent at completing the text when given an initial prompt, however, they are not ideal for NLP tasks where they need to follow instructions \cite{ouyang2022training}. Due to these advantages, we decided to utilize a fine-tuned (instruct) version of LLMs to reduce the manual effort of identifying concern-related app reviews.

\begin{algorithm*}
\caption{\textbf{RQ1}: NLI inference - Identifying best hypothesis and corresponding NLI model}\label{alg:cap}
\begin{algorithmic}[1]
\State \textbf{Input:} List of 31 generic hypotheses from \cite{harkous2022hark}, Heuristics \cite{harkous2022hark}, Newly defined domain-specific hypotheses and corresponding heuristics, and ground truth data from \cite{ebrahimi2022unsupervised}
\State \textbf{Output:} Best performing NLI model, Best of the two sets of hypotheses and Pseudo labeled corpus using best performing NLI model and best hypotheses
\State $generic\_hypotheses \gets $\text{list of 31 hypotheses from \cite{harkous2022hark}}, $heuristics \gets \text{set of heuristics from \cite{harkous2022hark}}$
\State $NLI\_models \gets \text{[Roberta-large-mnli, Nli-roberta-base, DeBERTa-v3-base-mnli-fever-anli, T5-base]}$
\State $domain\_specific\_hypotheses \gets \text{[domain specific hypotheses defined in Table \ref{tab:dhypo}]}$
\State $new\_heuristics \gets \text{[newly defined heuristics]}$, $dataset \gets \text{ground truth data from \cite{ebrahimi2022unsupervised}}$

\State $best\_NLI\_model = NLI\_models[0]$,  $best\_F1\_score = 0$
\For{$model \in NLI\_models$}
    \State $entailment\_scores = NLI\_Inference(model, generic\_hypotheses, dataset)$
    \State $nli\_annotated\_corpus = Apply\_Heuristics(entailment\_scores, heuristics)$
    \State $P, R, F1 = \text{\textit{Compute (Precision, Recall, and F1)}}$
    \If{$F1 > best\_F1\_score$}
        \State $\textbf{best\_NLI\_model} = model$
        \Comment {Determine the best performing NLI model on generic hypotheses and heuristics}
        \State $\textbf{best\_F1\_score} = F1$
    \EndIf
\EndFor

\State Next, use the best-performing NLI model on the domain-specific hypothesis
\State $entailement\_scores = NLI\_Inference(best\_NLI\_model, domain\_specific\_hypotheses, dataset)$
\State $nli\_annotated\_corpus = Apply\_Heuristics(entailment\_scores, new\_heuristics)$
\State $P, R, F1 = \text{\textit{Compute (Precision, Recall, and F1)}}$
\Comment{Metrics for best-performing NLI model with domain-specific hypotheses and new heuristics}
\State Next, determine which set of hypotheses is best performing by comparing F1 scores
\If {$F1 > best\_F1\_score$}
    \State $\textbf{best\_hypotheses} = domain\_specific\_hypotheses$
    \State $\textbf{best\_F1\_score} = F1$
\Else
    \State $\textbf{best\_hypotheses} = generic\_hypotheses$
\EndIf
\State $\textbf{pseudo\_labeled\_corpus} = labels(dataset, best\_NLI\_model, best\_hypotheses)$
\Comment{Corpus containing `maybe-privacy' and `maybe-not-privacy' labels}
\end{algorithmic}
\end{algorithm*}
\begin{figure*}[h]
    \centering
    \includegraphics[width=1\linewidth]{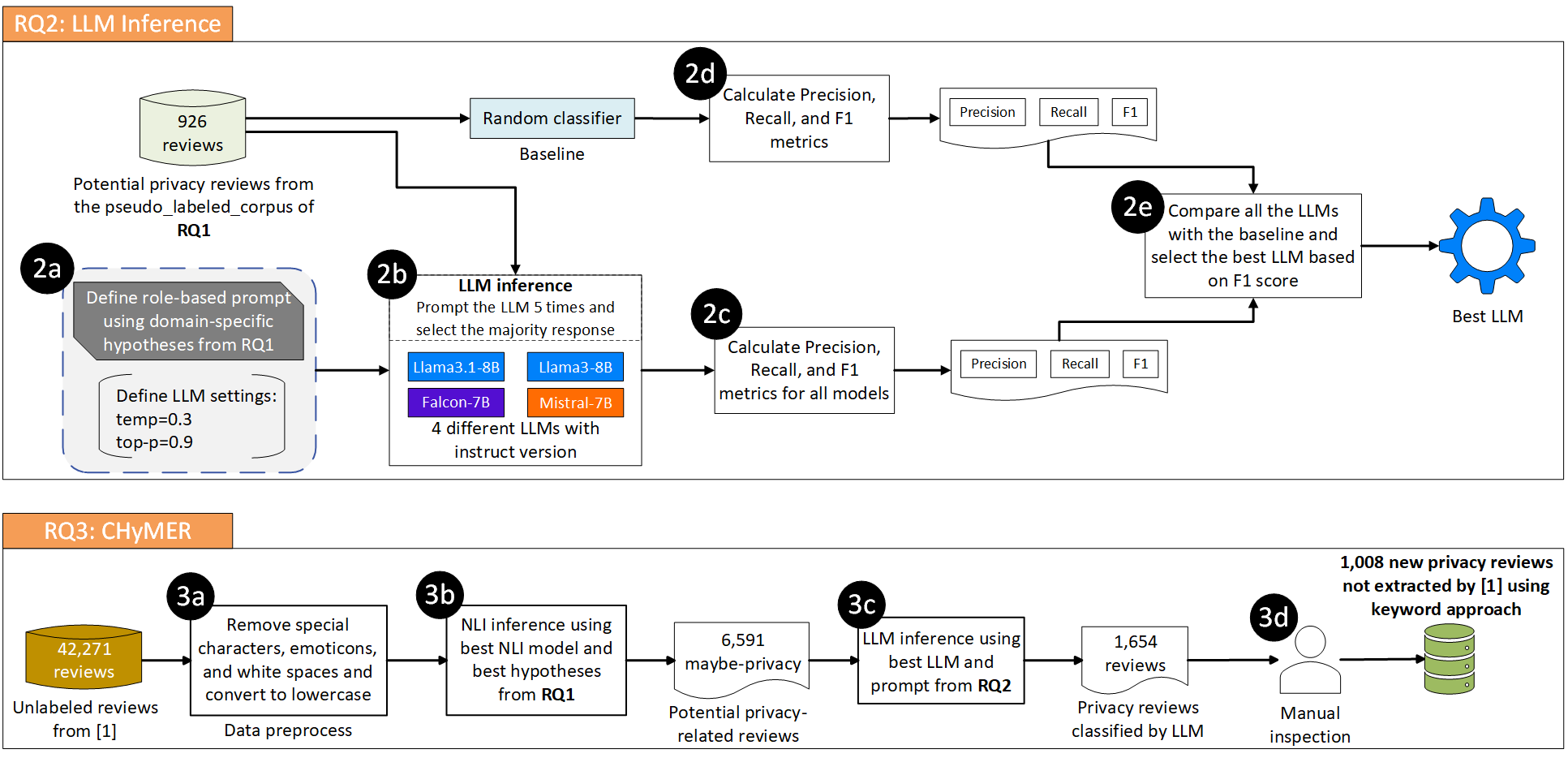}
    \caption{Overview of our methodology for LLM inference (RQ2) and extracting concern-related reviews using NLI+LLM (RQ3).}
    \label{fig:study}
\end{figure*}
\section{Dataset} \label{data}
\begin{table}[h]
    \centering
    \renewcommand{\arraystretch}{1.1}
    \caption{Statistics of the dataset used from \cite{ebrahimi2022unsupervised}.}
    \label{tab:stat}
    \begin{tabular}{l|l}
        \textbf{Number of apps} & 5 \\
        \hline
        \textbf{App category} & Health \& Fitness (MH) \\
        \hline
        \textbf{Total reviews} & 204,374 \\
        \hline
        \textbf{1-2 star rated reviews} & 43,647 \\
        \hline
        \textbf{Privacy labeled reviews} & 414 \\
        \hline
        \textbf{Non-privacy labeled reviews} & 962 \\
        \hline
        \textbf{Average number of words per review} & 33 \\
        \hline
        \textbf{Time range} & 2012-01-07 to 2021-10-06
    \end{tabular}
\end{table}
\begin{table}[h]
    \centering
    \renewcommand{\arraystretch}{1.1}
    \caption{Number of app reviews extracted for each app in MH domain.}
    \label{tab:apps}
    \begin{tabular}{p{2cm}|p{2cm}|p{3.5cm}}
        \textbf{App name} & \textbf{\# of Reviews} & \textbf{\# of 1-2 star rated reviews} \\
        \hline
        Calm & 106,181 & 22,983 \\
        Headspace & 78,989 & 16,376 \\
        Sanvelo & 8,554 & 698 \\
        Talkspace & 5,054 & 2,928 \\
        Shine & 5,596 & 662 \\
        \hline
        \textbf{Total} & 204,374 & 43,647 \\
    \end{tabular}
\end{table}
We utilize the ground truth data (manually validated), consisting of 1,376 privacy reviews from Ebrahimi et al. \cite{ebrahimi2022unsupervised} in this study. Table \ref{tab:stat} shows statistical information about the dataset. This particular dataset was developed through the application of keyword-matching filtering alongside manual inspection of over 204K reviews mined from the most widely used Mental Health (MH) applications available on the Google Play Store and Apple App Store. Although the raw dataset consisted of reviews from three application domains: MH, finance (investment), and food delivery; for this study, we exclusively focused on reviews pertinent to MH applications. Mainly because of a notable increase in the number of active users of MH applications as a consequence of the COVID-19 pandemic \cite{ebrahimi2022unsupervised} in the recent past. Individuals increasingly turn to these applications as a safer and more cost-effective means of addressing the psychological ramifications of social isolation, unemployment, and economic distress \cite{longyear2021can}.

To collect the app reviews from the MH domain, the authors identified the top 100 apps in the Health\&Fittness (MH) category on Google Play and the Apple App Store. Only the apps with 5,000 or more reviews were considered to include only popular and well-established apps. Additionally, physical health apps that did not explicitly support mental health were excluded. After examining the top 100 apps, five MH apps were selected for the analysis of reviews. For each of these apps, they collected all textual reviews available on the Apple App Store and Google Play using Python web scrapers. Overall, 204,374 reviews were collected. Table \ref{tab:apps} shows these reviews' distribution over apps.


Using the manual labeling method with a seed of {privacy, private, security} keywords, Ebrahimi et al. \cite{ebrahimi2022unsupervised} annotated 1,376 reviews with 1 (privacy-related) and 0 (non-privacy-related) labels. All these reviews contained keywords likely to indicate privacy concerns but only 414 reviews were privacy-related and 962 were non-privacy-related reviews. The data collection and labeling process are presented in detail in the study of Ebrahimi et al. \cite{ebrahimi2022unsupervised}.

In this study, we used the labeled dataset of 1,376 reviews to answer RQ1 and RQ2, and for RQ3 we used all 43,647 reviews rated with 1 or 2 stars (excluding the labeled reviews as they are part of the labeled sample of 1,376 reviews) as a source to extract the ethical-concern related app reviews further.

\section{Methodology} \label{std}

Algorithm \ref{alg:cap} and Figure \ref{fig:study} provide an overview of our research methodology. Our approach consists of three parts: (1) NLI-inference: we identify the best NLI model and the best set of hypotheses to extract potential privacy-related app reviews; (Algorithm \ref{alg:cap}) (2) LLM-inference: we then compare the performance of various LLaMA-like LLMs to classify potential reviews to privacy concerns and identify the best-performing LLM; (3) Finally, we combine NLI from RQ1 (NLI-inference) and LLM from RQ2 (LLM inference) to extract new privacy-related reviews (Figure \ref{fig:study}). 

We detailed the methods employed in each component in the following subsections.




\underline{\textbf{1) NLI Inference:}}
Algorithm \ref{alg:cap} describes the proposed NLI inference process. Using the existing 31 generic hypotheses (our hypotheses baseline) and the corresponding heuristics from Harkous et al. \cite{harkous2022hark}, and ground truth dataset from Ebrahim et al. \cite{ebrahimi2022unsupervised} \textbf{(lines 1-7)}, we first determine the best NLI model \textbf{(lines 8-16)} of the four chosen NLI models namely: Roberta-large-mnli, Nli-roberta-base, DeBERTa-v3-base-mnli-fever-anli and T5-base using Precision (P), Recall (R), and F1-score as measures.  We performed 1,376 (number of app reviews) * 31 (generic hypotheses) * 4 (number of models) = 170,624 inference operations at this stage.

Next, to determine the best hypotheses, we compare the performance of the generic hypotheses (baseline) with the newly defined domain-specific hypothesis and respective corresponding heuristics \textbf{(lines 17-27)}, which were determined manually. We performed 1,376 (number of app reviews) * 21 (domain-specific hypotheses) = 28,896 inference operations. In the end (\textbf{line 28}), we use the best NLI model and best hypotheses with their corresponding heuristics to create a pseudo-labeled corpus containing `maybe-privacy' and `maybe-not-privacy' labels. This pseudo-labeled corpus is further used for the evaluation of LLMs in RQ2.


\textbf{Generic hypotheses and corresponding heuristics (Baseline for RQ1):}
We use the generic privacy hypotheses and corresponding heuristics provided by Harkous et al. \cite{harkous2022hark} as a baseline for RQ1. Harkous et al. defined 31 generic hypotheses (Table \ref{tab:ghypo}) based on Solove's \cite{solove2005taxonomy} taxonomy of privacy violations and the taxonomy of privacy-enhancing technologies proposed by Wang and Kobsa \cite{wang2009privacy}. Further, they define the following heuristics where \textit{N\textsubscript{E}(i, t)} is the number of hypotheses receiving an entailment score above a threshold \textit{t} for review \textit{i}:

\begin{itemize}
    \item A review i is labeled as \textit{maybe-privacy} (potential privacy-related reviews) if \textit{N\textsubscript{E}(i, 0.8)$>$=1} or \textit{N\textsubscript{E}(i, 0.7)$>$=3} or \textit{N\textsubscript{E}(i, 0.6)$>$=5} or \textit{N\textsubscript{E}(i, 0.5)$>$=7}.
	\item A review i is labeled as \textit{maybe-not-privacy} if \textit{N\textsubscript{E}(i, 0.4)=0}. 
	\item Rest of the reviews are labeled as \textit{undetermined}.
\end{itemize}

\begin{table}[h]
    \centering
    \caption{Generic Privacy Concepts and Associated Hypotheses from \cite{harkous2022hark}}
    \label{tab:ghypo}
    \begin{tabular}{p{2cm}|p{6cm}}
        \hline
        \textbf{Privacy Concept} & \textbf{Hypotheses} \\
        \hline 
        \multicolumn{2}{l}{\textbf{Concepts from Solove’s Taxonomy \cite{solove2005taxonomy}}} \\
        \hline
        Surveillance & 1. The user is facing a data surveillance issue. \\
        Interrogation & 2. The user is forced to provide information. \\
        Aggregation & 3. Personal user information is collected from other sources. \\
        Insecurity & 4. The user is concerned about protecting their personal data. \\
        Identification & 5. A data anonymity topic is discussed. \\
        Secondary Use & 6. The user is concerned about the purposes of personal data access. \\
        Exclusion & 7. The user wants to correct their personal information. \\
        Breach of Confidentiality & 8. A breach of data confidentiality is discussed. \\
        Disclosure & 9. Personal data disclosure is discussed. \\
        Exposure & 10. The app exposes a private aspect of the user life. \\
        Increased Accessibility & 11. User’s data has been made accessible to public. \\
        Blackmail & 12. A data blackmailing issue is discussed. \\
        Appropriation & 13. User data is being exploited for other purposes. \\
        Distortion & 14. False data is presented about the user. \\
        Intrusion & 15. Unwanted intrusion to personal info is discussed. \\
        Decisional Interference & 16. Intrusion by the government to the user’s life is discussed. \\
        \hline
        \multicolumn{2}{l}{\textbf{Concepts from Wang and Kobsa’s Taxonomy \cite{wang2009privacy}}} \\
        \hline
        Notice/Awareness & 17. Opting out from personal data collection is discussed. \\
        Data Minimization & 18. More access than needed is required. \\
        Purpose Specification & 19. The reason for data access is not provided. \\
        Collection Limitation & 20. Too much personal data is collected. \\
        Use Limitation & 21. The data is being used for unexpected purposes. \\
        Onward Transfer & 22. Data sharing with third parties is discussed. \\
        Choice/Consent & 23. User choice for personal data collection is discussed. \\
        & 24. User did not allow access to their personal data. \\
        \hline
        \multicolumn{2}{l}{\textbf{Generic Privacy Concepts}} \\
        \hline
        \multirow{1}{2cm}{Generic Privacy Issues} & 25. A data privacy topic is discussed. \\
        & 26. Protecting user’s personal data is discussed. \\
        & 27. This is about a privacy feature. \\
        & 28. The user is facing a privacy issue. \\
        \multirow{1}{2cm}{Positive Privacy Issues} & 29. The user likes that data privacy is provided. \\
        & 30. The user wants privacy. \\
        & 31. The app has privacy features.
    \end{tabular}
\end{table}

\textbf{Determining domain-specific hypotheses and corresponding heuristics (Our approach):}
We manually define the domain-specific privacy hypotheses based on the Mental Health (domain-specific) privacy concepts provided by Iwaya et al. \cite{iwaya2023privacy} in their exploration of MH applications development. For each concept, following the method from  Harkous et al. \cite{harkous2022hark}, we came up with one or more hypotheses. For example, for the ``\textit{Non-repudiation}" concept we defined two hypotheses: ``\textit{User is unable to deny their online actions.}" and ``\textit{User is concerned about the permanent storage of their digital transactions.}". In total, we defined 21 domain-specific privacy hypotheses as shown in Table \ref{tab:dhypo}.

Further, similar to previous studies \cite{harkous2022hark, duvsek2020evaluating}, we define the following heuristics to sample our reviews based on the entailment scores and label the reviews with \textit{maybe-privacy}  and \textit{maybe-not-privacy} labels.

\begin{itemize}
    \item A review i is labeled as \textit{maybe-privacy} if \textit{N\textsubscript{E}(i, 0.85)$>$=1} or \textit{N\textsubscript{E}(i, 0.75)$>$=3} or \textit{N\textsubscript{E}(i, 0.7)$>$=5}.
	\item Rest reviews are labeled as \textit{maybe-not-privacy}.
\end{itemize}

The intuition behind defining this heuristic is to select the most potentially privacy-related reviews with high confidence by minimizing the FP (0-labeled reviews annotated as `maybe-privacy') and FN (1-labeled reviews annotated as `maybe-not-privacy').

\begin{table}[h]
    \centering
    \caption{MH Domain-specific Privacy Concepts \cite{iwaya2023privacy} and Associated Hypotheses}
    \label{tab:dhypo}
    \begin{tabular}{p{1.5cm}|p{6.5cm}}
        \hline
        \textbf{Privacy Concept} & \textbf{Hypotheses} \\
        \hline 
        \multicolumn{2}{l}{\textbf{Concepts from Iwaya et al’s Taxonomy \cite{iwaya2023privacy}}} \\
        \hline
        \multirow{1}{1.5cm}{Linkability} & 1. User data being linked across different services. \\
        & 2. Online user activities from various platforms can be connected. \\
        & 3. Personal user information is collected from other sources. \\
        \multirow{1}{1.5cm}{Identifiability} & 4. Anonymized user data could be used to reveal their identity. \\
        & 5. Unique digital user data could lead to personal identification. \\
        \multirow{1}{1.5cm}{Non-repudiation} & 6. User is unable to deny their online actions. \\
        & 7. User is concerned about the permanent storage of their digital transactions. \\
        \multirow{1}{1.5cm}{Detectability} & 8. User is concerned about others detecting their use of sensitive online services. \\
        & 9. User presence on certain platforms could be discovered from anonymized data. \\
        \multirow{1}{1.5cm}{Disclosure of information} & 10. User device's communication patterns reveal private information. \\
        & 11. User device's communication patterns reveal private information. \\
        & 12. The app exposes a private aspect of the user life. \\
        \multirow{1}{1.5cm}{Unawareness} & 13. Unauthorized access to user's private information. \\
        & 14. The user is not aware of how and why their data is being collected, processed, stored, and shared. \\
        \multirow{1}{1.5cm}{Non-compliance} & 15. The user is concerned about the processing or storing of their personal data against regulations or privacy policies. \\
        & 16. User data is being exploited for other purposes. \\
        & 17. Data sharing with third parties is discussed. \\
        \hline
        \multicolumn{2}{l}{\textbf{Additional Privacy Concepts}} \\
        \hline
        \multirow{1}{1.5cm}{General Privacy Issues} & 18 The user is facing a privacy issue. \\
        & 19. The user is concerned about protecting their personal data. \\
        & 20. A data anonymity topic is discussed. \\
        & 21. A data privacy topic is discussed.
    \end{tabular}
\end{table}

\textbf{NLI Models:} We perform inference with four different NLI models. These models are chosen due to their state-of-the-art NLI performance and easy availability on the HuggingFace platform \cite{wolf2019huggingface}. Additionally, these models are fine-tuned and pre-trained for the NLI task using state-of-the-art NLI datasets namely: \\
- Multi-Genre Natural Language Inference (MNLI) \cite{williams2017broad} (433k sentence pairs) \\
- Adversarial Natural Language Inference (ANLI) \cite{nie2019adversarial} (169k sentence pairs) \\
- Stanford Natural Language Inference (SNLI) \cite{bowman2015large} (570k sentence pairs) \\
- Question Answer NLI (QNLI) \cite{rajpurkar2016squad} (116k sentence pairs) \\
- Fact Extraction and VERification NLI (FeverNLI) \cite{thorne2019fever2} (185k sentence pairs)

\noindent The four chosen NLI models are as follows:
\begin{itemize}[leftmargin=*]
    \item \textbf{Roberta-large-mnli}: is the RoBERTa large model \cite{liu2019roberta} fine-tuned on the MNLI dataset. The model is pre-trained on English language text using a masked language modeling (MLM) objective.
    \item \textbf{Nli-roberta-base}: is the RoBERTa base model \cite{liu2019roberta} fine-tuned on the MNLI and SNLI datasets using Sentence Transformers Cross-Encoder class \cite{reimers-2019-sentence-bert}.
    \item \textbf{DeBERTa-v3-base-mnli-fever-anli}: is the DeBERTa-v3 base model \cite{he2020deberta} fine-tuned on the MNLI, FeverNLI, and ANLI datasets.
    \item \textbf{T5-base}: is the vanilla T5 model \cite{2020t5} readily fine-tuned on the MNLI and QNLI datasets.
\end{itemize}

\noindent To implement these models, we use the transformers library from HuggingFace \cite{wolf2019huggingface} with their respective tokenizers. 

\underline{\textbf{2) LLM Inference:}}
Figure \ref{fig:study} outlines the proposed LLM inference process. We use 926 potential privacy-related (annotated as `maybe-privacy') reviews from the pseudo-labeled corpus created in the NLI inference component. First, we design the prompt and configure the LLM settings as shown in step \circled{2a}. Next, we choose four different LLaMA-like LLMs of the instruct version, perform the inference operation (step \circled{2b}), and calculate the P, R, and F1-score from the results (step \circled{2c}). Next, we calculate these metrics for the baseline Random Classifier (RC) (step \circled{2d}) and compare the results of LLMs with the baseline to select the best-performing LLM (step \circled{2e}).

\textbf{Choice of LLaMA-like LLMs:}
To make our study replicable and more accessible we choose four different open-source LLMs of the instruct versions that have state-of-the-art performance and are readily available through the transformers library of HuggingFace \cite{wolf2019huggingface}.

\begin{itemize}[leftmargin=*]
    \item \textbf{meta-llama/Llama-3.1-8B-Instruct} \cite{dubey2024llama}: is the LLaMA 3.1 instruction-tuned text-only auto-regressive language model that uses an optimized transformer architecture. The tuned versions use supervised fine-tuning (SFT) and reinforcement learning with human feedback (RLHF) to align with human preferences for helpfulness and safety.
	\item \textbf{meta-llama/Meta-Llama-3-8B-Instruct} \cite{dubey2024llama}: is the LLaMA 3 instruction-tuned text-only auto-regressive language model.
	\item \textbf{tiiuae/falcon-7b-instruct} \cite{almazrouei2023falcon}: is the Falcon-7B base model finetuned on a mixture of chat/instruct datasets.
	\item \textbf{mistralai/Mistral-7B-Instruct-v0.3} \cite{jiang2023mistral}: is an instruct fine-tuned version of the Mistral-7B-v0.3 base model.
\end{itemize}

\textbf{Random Classifier (Baseline for RQ2)}
Similar studies on app review classification have compared their approaches to either the current state-of-the-art or a baseline RC \cite{alomar2021finding, obie2022violation}. Hence we compare our best-performing LLM with a baseline RC only since there is no current LLaMA-like LLM-based state-of-the-art in classifying ethical concern-related app reviews, similar to what recent works have done \cite{alomar2021finding, obie2022violation}.

\textbf{Prompt Design and LLM Settings (Our approach):}

\begin{figure}[b]
    \centering\includegraphics[width=1\linewidth]{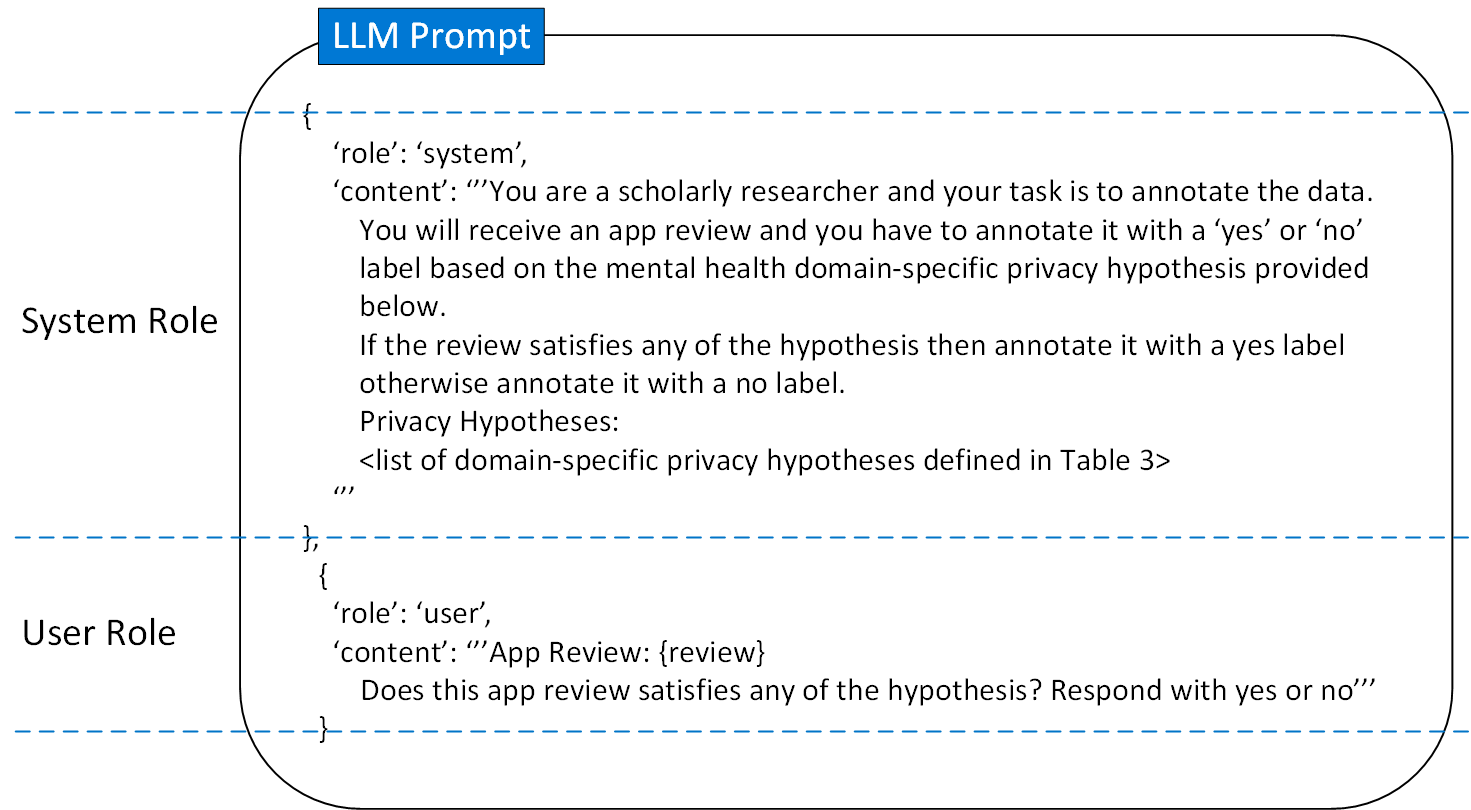}
    \caption{Role-based prompt designed using domain-specific privacy hypotheses defined in Table \ref{tab:dhypo}. The \textit{system} role is used to give the instructions to the LLM and the \textit{user} role is to provide the app review and get the response.}
    \label{fig:prompt}
\end{figure}

In line with previous work \cite{touvron2023llama, zhang2023revisiting}, we build prompts for the LLaMA-like LLMs to experiment with zero-shot setting and we follow the guidelines provided by \cite{chen2023unleashing} to design the prompt and configure LLM settings.

The LLM settings namely, the temperature and the top-\textit{p}, play a crucial role in the generation of responses \cite{chen2023unleashing}. The temperature parameter controls the randomness of the generated output: a lower temperature leads to more deterministic outputs \cite{chen2023unleashing}. The top-p parameter, on the other hand, controls the nucleus sampling, which is a method to add randomness to the model’s output \cite{chen2023unleashing}. Adjusting these parameters can significantly affect the quality and diversity of the model’s responses, making them essential tools in prompt engineering \cite{chen2023unleashing}. Thus, we set the temperature value to 0.3 and the top-p parameter to 0.9 to get deterministic, coherent, and contextually relevant responses. Additionally, we prompt the model five times and select the majority response to overcome the inherent variability in the model’s responses and increase the chances of obtaining a more deterministic output \cite{chen2023unleashing}.

To design the prompt, we followed the role prompting technique by defining clear and precise instructions for each role. This technique involves giving the model a specific role, such as a helpful assistant or an expert \cite{chen2023unleashing}. It can be particularly effective in guiding the model’s responses and ensuring that they align with the desired output \cite{chen2023unleashing}. In our design, we defined two roles system and user, for the model. In the system role, we give the instructions to the model to act as a scholarly researcher and annotate the given app review with yes/no labels, and in the user role, we provide the model with the app reviews and get the response. Additionally, in the system role instructions we add the domain-specific hypotheses based on which LLM is instructed to respond. Figure \ref{fig:prompt} shows the detailed prompt structure.

\underline{\textbf{3) NLI+LLM}:} 
Figure \ref{fig:study} shows the complete flow of our hybrid proposed (NLI with LLM) approach. We utilize the set of 42,271 unlabeled app reviews from \cite{ebrahimi2022unsupervised}. First, we preprocess all the reviews to remove all the special characters, emoticons, and white spaces, and convert them to lowercase (step \circled{3a}). Next, we perform the NLI inference using the best NLI model and the best set of hypotheses from RQ1 to filter out the non-relevant reviews and get a set of potential privacy-related (maybe-privacy) reviews (step \circled{3b}). Then, we perform LLM inference using the best LLM and prompt from RQ2 to get the relevant privacy-related reviews (step \circled{3c}). At the end (step \circled{3d}), we perform the manual inspection to filter out the wrongly classified reviews further as LLMs are not 100\% accurate.

\textbf{Keyword-based approach (baseline for RQ3)} 
For the evaluation of NLI+LLM results, we use the keyword-based approach as our baseline. Here, we try to show that NLI+LLM can extract concern-related reviews that were missed by the baseline keyword-based approach utilized by \cite{ebrahimi2022unsupervised}.

\textbf{Manual inspection setup:}
The inspection task is to identify whether the extracted reviews contain any privacy concerns and it is carried out to create a ground truth dataset for the research community to conduct further studies. Four annotators including the first author and 3 graduate students from our research lab conducted this task. The first author analyzes all the reviews while the others inspect one-third of the sample such that each review is inspected at least twice. To prevent exhaustion, we perform this process in a 7-day timeframe. 

To ensure the understanding of the task and the definitions for privacy and non-privacy labels, we create the labeling instructions (available in our replication package) and we base our analysis on the privacy concepts provided by Iwaya et al. \cite{iwaya2023privacy} to label the reviews. After the manual inspection, we cross-check the findings of the manual classification. For every disagreement, a third annotator is requested to break the tie. In total, 137 reviews had to be further analyzed by another annotator. To determine the extent to which the annotators agreed upon the classifications, we use Cohen’s Kappa coefficient \cite{cohen1960coefficient}. We acquired a degree of agreement of 0.82. According to Fleiss et al. \cite{fleiss2013statistical}, this agreement value is nearly perfect agreement (i.e., 0.80 - 1.00). In that sense, the resultant sample results from a process for which all annotators agreed on 100\%.

\underline{\textbf{4) Evaluation measures:}}
To evaluate the performance of NLI and LLM inference operations, we employ the measures of P, R, and F1-score, similar to the previous studies \cite{duvsek2020evaluating, obie2022violation}. The F1-score (\(F1 = \frac{2*P*R}{P+R}\)) corresponds to the harmonic mean of P (\(P = \frac{TP}{TP+FP}\)) and R (\(R = \frac{TP}{TP+FN}\)), where P is the number of correct predictions out of all the input sample and R is number of positive predictions that was observed in the actual class.

In case of NLI evaluation, True Positives (TP) refers to the number of 1-labeled reviews annotated with a `maybe-privacy' label, True Negatives (TN) refers to the number of 0-labeled reviews annotated with `maybe-not-privacy' and `undetermined' labels, FP refers to the number of 0-labeled reviews annotated with `maybe-privacy' label and FN refers to the number of 1-labeled reviews annotated with the `maybe-not-privacy' or `undetermined' labels.

In case of LLM evaluation, TP refers to the number of 1-labeled reviews classified with the `yes' label, TN refers to the number of 0-labeled reviews classified with the `no' label, FP refers to the number of 0-labeled reviews classified with the `yes' label and FN refers to the number of 1-labeled reviews classified with the `no' label.


\underline{\textbf{5) Computational resources:}}
The experiments are conducted on an NVIDIA GeForce RTX 4090 GPU of 40 GB RAM, NVIDIA-SMI driver version 546.09, and a 24-core CPU setup. We implement our models using Python 3.12 with CUDA version 11.8 and HuggingFace Transformers version 4.44.1. We use NumPy and Pandas for linear algebra operations. These resources enhance our experiments' efficiency and scalability and ensure our study's reproducibility in comparable high-performance computing environments.

\section{Results} \label{str}

This section presents and discusses the results of our investigation. For each RQ, we present the results of the analysis, and we discuss the findings.

\textbf{RQ1 - NLI (Domain-specific hypotheses vs Generic hypotheses (Baseline)): }
First, we evaluate four NLI models using the baseline generic privacy hypotheses and select the best NLI model based on the highest F1-score. Table \ref{tab:rq1} shows the inference results for the generic hypotheses. DeBERTa-v3-base-mnli-fever-anli is the best-performing model with the highest F1-score of 0.5. It can be observed that all the models have low P values as NLI identifies the high number of FP. DeBERTa-v3-base-mnli-fever-anli annotated 1130/1376 reviews as `maybe-privacy' along with achieving the goal of minimizing FP and FN. It annotated only 25 1-labeled reviews as `maybe-not-privacy' and 741 0-labeled reviews as `maybe-privacy'. Additionally, it can be noted that the resultant metrics of the T5-base model are 0 as it has 0 TP and FP.

Next, we compare the inference results of the domain-specific hypotheses with generic hypotheses using the best-performing DeBERTa-v3-base-mnli-fever-anli model. Table \ref{tab:rq1_2} shows the findings indicating that domain-specific hypotheses yield better results as compared to generic hypotheses. We achieved an F1-score of 0.54 with domain-specific hypotheses which shows an improvement of 1.08 times as compared to the generic hypotheses. This improvement is promising in terms of FP as we identified only 568 FP in the case of domain-specific hypotheses whereas this count was comparatively higher (741) for generic hypotheses.

\begin{mdframed}[backgroundcolor=lightgray]
Summary of RQ1: The DeBERTa-v3-base-mnli-fever-anli NLI model with domain-specific hypotheses and corresponding heuristics performs best in extracting potential concern-related app reviews. It achieves an F1-score of 0.54 and 1.08 times improvement compared to the baseline generic hypotheses.
\end{mdframed}

\begin{table}[h]
    \centering
     \renewcommand{\arraystretch}{1.2}
    \caption{Results of NLI inference using the baseline generic hypotheses and corresponding heuristics.}
    \label{tab:rq1}
    \begin{tabular}{p{4.3cm}|p{.5cm}p{.5cm}p{.5cm}}
  
        \textbf{Model} & \textbf{P} & \textbf{R} & \textbf{F\textsubscript{1}} \\ \hline
 
        Roberta-large-mnli & 0.35 & 0.8 & 0.49 \\
  
        \textbf{DeBERTa-v3-base-mnli-fever-anli} & 0.34 & 0.93 & \textbf{0.50} \\
    
        T5-base & 0 & 0 & 0 \\
 
        Nli-roberta-base & 0.32 & 0.96 & 0.48 \\
 
    \end{tabular}
\end{table}

\begin{table}[h]
    \centering
        \renewcommand{\arraystretch}{1.2}
    \caption{Comparison of NLI inference using domain-specific hypotheses and baseline generic hypotheses with the best NLI model.}
    \label{tab:rq1_2}
    \begin{tabular}{p{2.3cm}|p{.5cm}p{.5cm}p{.5cm}|p{2.7cm}}
     
        \textbf{Hypotheses} & \textbf{P} & \textbf{R} & \textbf{F\textsubscript{1}} & \textbf{Improvement on F1} \\ \hline
 
        Generic (Baseline) & 0.34 & 0.93 & 0.50 & - \\
 
        Domain-specific & 0.39 & 0.86 & \textbf{0.54} & \textbf{1.08x}   
    \end{tabular}
\end{table}

\textbf{RQ2 - LLaMA-like LLM vs RC (Baseline): }
In LLM inference, we use the 926 `maybe-privacy' reviews from the pseudo-labeled corpus of the DeBERTa-v3-base-mnli-fever-anli model with domain-specific hypotheses. For our baseline, we use the statistics of our dataset to calculate the metrics. The precision of the baseline RC is computed by dividing the number of privacy reviews by the total number of reviews (i.e., \(\frac{358}{926}=0.38\)). Regarding recall, there is only a 50\% probability for a review to be classified as a privacy review since there are two possible
classifications available. Finally, the F1-measure of baseline RC is calculated as \(2*\frac{0.38*0.5}{0.38+0.5}=0.43\).

Table \ref{tab:rq2} shows the P, R, and F1-score of all the LLMs and the baseline RC along with the improvement in the F1-score of LLMs as compared to RC. These results highlight that the Llama3.1-8B-Instruct model achieved the best performance with an F1-score of 0.81 and an improvement of 1.86 times as compared to the RC. While the Llama-3-8B-Instruct model is the second best performing model with an F1-score of 0.69, the falcon-7b-instruct model achieved the highest recall of 0.95.

\begin{mdframed}[backgroundcolor=lightgray]
Summary of RQ2: Llama3.1-8B-Instruct LLM shows the best performance for extracting concern-related reviews from the potential set of reviews with an F1=0.81 and 1.86 times improvement as compared to the baseline RC.
\end{mdframed}

\begin{table}[h]
    \centering
    \caption{LLM inference results on the dataset of 926 \textit{maybe-privacy} reviews and their comparison with the baseline RC. The last column shows the improvement on F1-score as compared to the baseline}
    \label{tab:rq2}
    \renewcommand{\arraystretch}{1.1}
    \begin{tabular}{p{2.8cm}|p{.5cm}p{.5cm}p{.5cm}|p{2cm}}
        \textbf{Model} & \textbf{P} & \textbf{R} & \textbf{F1} & \textbf{F1 improvement} \\
        \hline
        RC \textbf{(Baseline)} & 0.38 & 0.5 & 0.43 & - \\
     
        Llama-3.1-8B-Instruct & 0.72 & 0.92 & \textbf{0.81} & \textbf{1.86x} \\
      
        Llama-3-8B-Instruct & 0.59 & 0.83 & 0.69 & 1.6x \\
 
        Falcon-7b-instruct & 0.4 & 0.95 & 0.57 & 1.3x \\
 
        Mistral-7B-Instruct-v0.3 & 0.36 & 0.089 & 0.14 & 0.33x \\
 
    \end{tabular}
\end{table}

\textbf{RQ3 -  NLI+LLM vs Keyword-matching (Baseline): }
To extract the new set of privacy-related reviews from the dataset of 42,271 unlabeled reviews, we use the DeBERTa-v3-base-mnli-fever-anli model (best-performing NLI model) with the domain-specific hypotheses from RQ1 and Llama3.1-8B-Instruct LLM (best-performing LLM) with the prompt from RQ2. After data preprocessing, we executed the NLI inference and identified 6,591 `maybe-privacy’ reviews. These reviews were then used in LLM inference operation, and 1,654 reviews were further labeled as `yes' by the LLM indicating the privacy-related reviews. After this, we performed the manual inspection and created a dataset of 1,008 privacy-related reviews that were not extracted by the previous study \cite{ebrahimi2022unsupervised} using keyword-based filtering. We show a few of the reviews below and make the whole dataset publicly available.

\begin{mdframed}
\textbf{Review 1}: ``Don't bait people in to take their information and sell it and add them to your mailing list then force a paywall to use the app" \\
\textbf{Review 2}: ``How are you different from any other app now that is interested in our user patterns over our mental health?" \\
\textbf{Review 3}: ``This app has data trackers don t trust any app with your wellbeing that is sending your behavior data to multiple third parties"
\end{mdframed}

All these reviews mention privacy concerns but they do not contain any predefined set of keywords and are also specifically related to MH domain. This shows the importance of using NLI (with domain-specific hypotheses) and LLM to extract the ethical concern-related app reviews.

\begin{mdframed}[backgroundcolor=lightgray]
Summary of RQ3: 1,008 new privacy-related reviews were extracted using the best-performing NLI model with domain-specific hypotheses and the best LLM.
\end{mdframed}

\section{Threats to Validity} \label{threats}
The study presented in this paper has several limitations that could potentially limit the validity of the results.

\noindent \textbf{Construct threats}:
The potential threat to the construct validity of our study is related to the appropriateness of the study dataset and our manually created dataset. Developing a dataset is a tedious job and also subject to reader bias. We mitigated this risk by choosing a dataset of privacy reviews that were previously identified and validated through manual inspection by Ebrahimi et al. \cite{ebrahimi2022unsupervised}. Additionally, for curating a new dataset we employed a methodological approach for manual inspection, including four annotators to mitigate the risk of an individual bias.

Further, we utilize four NLI models and four LLMs with three evaluation metrics. Hence, we accept that applying other models to our dataset may lead to different results. The metrics P, R, and F1-score used in this study are widely applied and suggested to evaluate such models in SE.

\noindent\textbf{Internal threats}: 
The process of defining domain-specific privacy hypotheses and corresponding heuristics, and designing the prompt for LLMs may introduce some threats to the internal validity of our study. We used the technique suggested by previous studies to mitigate such threats. Similar to \cite{harkous2022hark}, we defined privacy hypotheses based on the widely used MH domain privacy taxonomies \cite{iwaya2023privacy}. Additionally, we followed the approach of \cite{harkous2022hark, duvsek2020evaluating} to define our corresponding heuristics. To design the prompt for LLMs we followed the guidelines provided by \cite{chen2023unleashing}.

Other potential threats to internal validity may emerge from the analysis being limited to reviews with only 1 and 2-star ratings. This limitation may have resulted in the omission of certain valuable reviews from the dataset. Nevertheless, recent research has indicated that reviews related to ethical concerns, particularly those related to privacy, frequently correlate with lower star ratings \cite{ebrahimi2022unsupervised}. Consequently, the exclusion of reviews with higher ratings is improbable to result in the neglect of significant concerns.

\noindent\textbf{External threats}: 
The main threat to the external validity of our study stems from the fact that only the MH domain reviews were considered from three application domain reviews provided by \cite{ebrahimi2022unsupervised}. Due to this, we acknowledge that our methodology could produce different results if applied to different domains and, our findings may not necessarily generalize to data from other app domains, and platforms other than Google Play Store and Apple App Store. Finally, as a further limitation to the external validity, we acknowledge that the choice of focusing on open-source LLMs limits the generalizability of our findings. We advocate in favor of future replications, including consideration of other – possibly enterprise or closed-source – models, such as GPT-4 by OpenAI.

\section{Conclusion and Future Work} \label{conclude}

We present an NLI+LLM-based approach that enables developers to proficiently discern ethical concerns associated with their applications and enhance them towards being more trustworthy and responsible by leveraging user feedback. Our objective is to foster sustainable change by integrating a model within the software development lifecycle of developers, while simultaneously elevating awareness regarding pre-existing ethical issues that obstruct the usability of mobile applications, which represents an urgent necessity \cite{sorathiya2024towards}.

While formulating our methodology, we undertook a comprehensive evaluation encompassing three distinct phases to address our three RQs and demonstrate the effectiveness of our approach. In the first phase, we evaluated four different NLI models to extract potential privacy reviews and compared the results of domain-specific privacy hypotheses with the generic privacy hypotheses. Our results showed that the DeBERTa-v3-base-mnli-fever-anli NLI model with domain-specific privacy hypotheses offered the best performance in extracting potential concern-related app reviews.

In the second phase of our analysis, we evaluated four LLaMA-like LLMs to classify concern-related reviews from the set of potential reviews. Our analysis showed that the Llama-3.1-8B-Instruct LLM was the best-performing model with an F1-score of 0.81. In the final phase of our analysis, we used NLI+LLM to extract new 1,008 privacy-related reviews from the dataset that were not extracted by the previous study using a keyword-based approach.

In future work, we intend to (i) leverage topic modeling to automatically identify the main topics addressed by users in concern-related reviews; (ii) create a user-friendly and interactive tool for developers to extract concern-related reviews and summarize them easily; (iii) automatically extract requirements from the concern-related reviews which can be directly addressed and implemented in the development phase; Moreover (iv) devise an interactive guide in which practitioners can explore concern-related topics and navigate through relevant reviews to understand the evidence for each recommendation.

\noindent\textbf{Data Availability:} All the data for this study is available here\footnote{\url{https://github.com/AakashSorathiya/CHyMER}}.

\bibliographystyle{IEEEtran}
\bibliography{short_ref}

\begin{thebibliography}{10}
\bibitem{gillespie_are_2019}
E.~Gillespie, ``Are you being scanned? how facial recognition technology follows you, even as you shop.''
\end{thebibliography}

\end{document}